\pdfoutput=1

\documentclass[11pt]{article}

\usepackage{ACL2023}

\usepackage{times}
\usepackage{latexsym}
\usepackage{graphicx}
\usepackage{amsfonts}
\usepackage{booktabs}
\usepackage{multirow}
\usepackage{makecell}
\usepackage{amsmath}
\usepackage{subfigure}

\usepackage[T1]{fontenc}

\usepackage[utf8]{inputenc}

\usepackage{microtype}

\usepackage{inconsolata}

\title{Understanding and Mitigating Gender Bias in LLMs \\ via Interpretable Neuron Editing}

\author{Zeping Yu \quad Sophia Ananiadou\\
  Department of Computer Science, National Centre for Text Mining \\
  The University of Manchester  \\
  \texttt{\{zeping.yu@postgrad. sophia.ananiadou@\}manchester.ac.uk}}

\begin{document}
\maketitle
\begin{abstract}
Large language models (LLMs) often exhibit gender bias, posing challenges for their safe deployment. Existing methods to mitigate bias lack a comprehensive understanding of its mechanisms or compromise the model’s core capabilities. To address these issues, we propose the CommonWords dataset, to systematically evaluate gender bias in LLMs. Our analysis reveals pervasive bias across models and identifies specific neuron circuits, including ``gender neurons'' and ``general neurons,'' responsible for this behavior. Notably, editing even a small number of general neurons can disrupt the model’s overall capabilities due to hierarchical neuron interactions. Based on these insights, we propose an interpretable neuron editing method that combines logit-based and causal-based strategies to selectively target biased neurons. Experiments on five LLMs demonstrate that our method effectively reduces gender bias while preserving the model’s original capabilities, outperforming existing fine-tuning and editing approaches. Our findings contribute a novel dataset, a detailed analysis of bias mechanisms, and a practical solution for mitigating gender bias in LLMs.
\end{abstract}

\section{Introduction}
Transformer-based \citep{vaswani2017attention} large language models (LLMs) \citep{brown2020language,ouyang2022training,chowdhery2023palm} have achieved remarkable breakthroughs and are widely applied in various NLP and multimodal tasks. While LLMs acquire powerful capabilities such as factual knowledge \cite{sun2023head}, reasoning \cite{wei2022chain}, and arithmetic ability \cite{yuan2023well} from large-scale corpora, they also learn undesirable gender bias \cite{ranaldi2023trip,o2024gender}. If left unchecked, LLMs may reproduce or even amplify this bias, leading to negative impacts in real-world applications. Therefore, reducing gender bias has become one of the most critical challenges in deploying LLMs responsibly.

Many studies \cite{zhao2018gender,webster2020measuring,pant2022incorporating,yang2023adept,ranaldi2023trip} have made progress in mitigating gender bias, but two major challenges remain. First, the storage and mechanisms underlying gender bias in LLMs are still not understood. Previous studies \cite{dai2021knowledge,geva2022transformer,yu2024interpreting} suggest that neurons are the fundamental units responsible for storing knowledge and computational operations in LLMs. If we could pinpoint the neurons responsible for gender bias, targeted editing of these neurons could effectively mitigate the bias. However, neuron-level research on gender bias in LLMs is limited, leading to an insufficient understanding of its mechanism and storage location. Second, current bias reduction techniques often overlook their effects on the model's original capabilities. Previous studies have shown that methods such as fine-tuning or model editing can disrupt the model’s performance on other tasks \cite{kirkpatrick2017overcoming,ramasesh2021effect,luo2023empirical,yang2024butterfly,gu2024model}. If these impacts are significant, removing gender bias may harm overall performance.

\begin{figure}[htb]
\vspace{-10pt}
\begin{center}
\centerline{\includegraphics[width=\columnwidth]{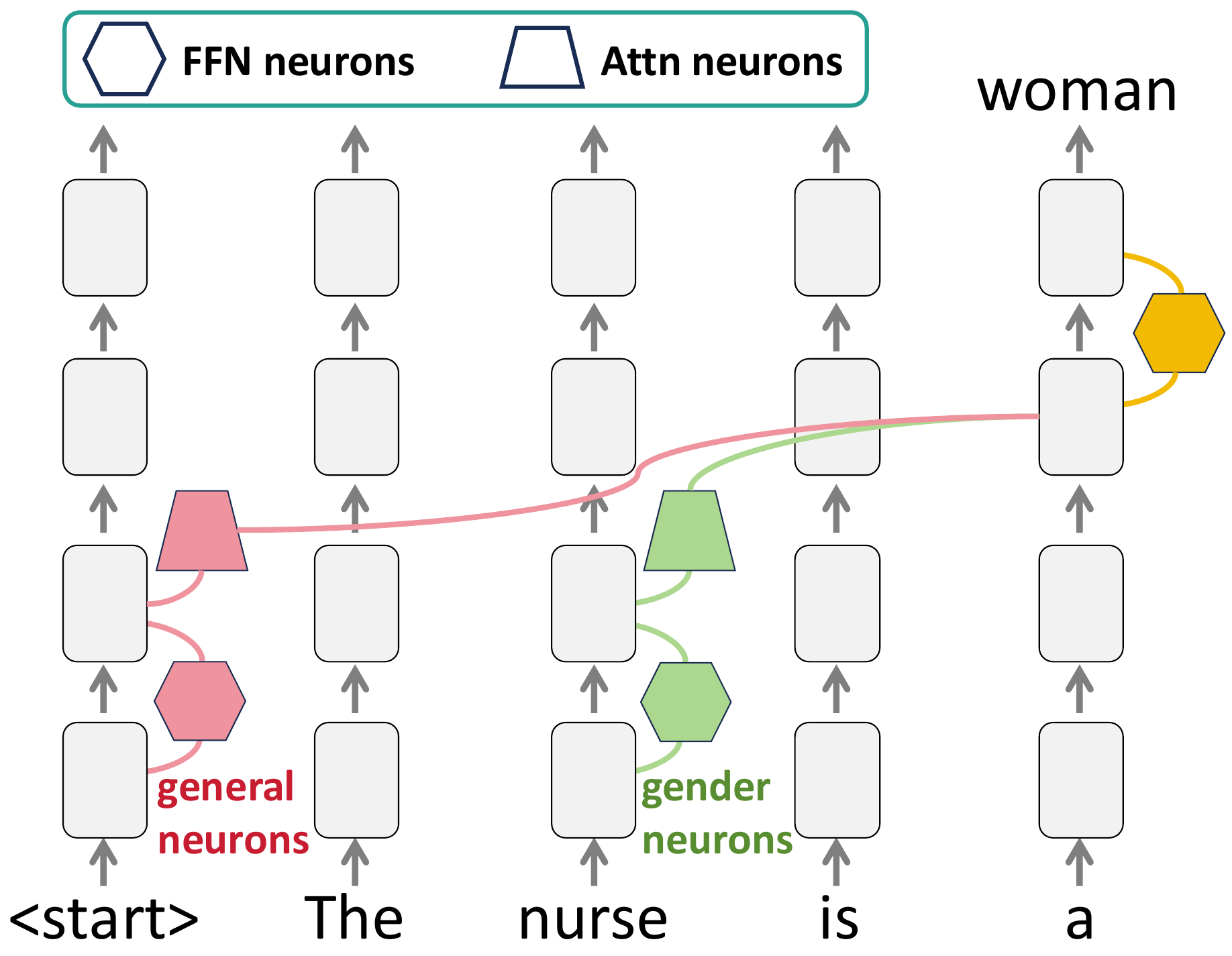}}
\caption{The neuron-level information flow of sentence ``The nurse is a'' -> ``woman''. The <start> token activates ``general neurons'' and the word ``nurse'' activates ``gender neurons'' on their residual streams. These information propagate through attention neurons and are transferred to the final position, ultimately contributing to the prediction of ``woman.''}
\label{figure1}
\end{center}
\vspace{-10pt}
\end{figure}

Addressing these challenges requires a deeper understanding of the neuron-level storage and information flow of gender bias, as well as strategies to mitigate bias while preserving the model’s core capabilities. Our approach addresses these challenges as follows. First, we introduce a new dataset, CommonWords, which consists of five categories of common words: traits, actions, professions, colors, and hobbies, with 100 words in each category. Using this dataset, we evaluate the gender preferences of five LLMs and observe that gender bias is pervasive across all models. Then, we analyze the neuron-level information flow to investigate the mechanisms behind specific instances of gender bias. We identify two distinct neuron circuits involved in gender bias, as shown in Figure 1. On one hand, stereotypical words trigger ``gender neurons'' in shallow layers, whose coefficients have opposite signs depending on different words. These activations propagate to higher-layer attention neurons and FFN neurons, influencing gender-specific predictions. On the other hand, the <start> token activates ``general neurons,'' leading to enhance the probability of common words. We further find that editing just two ``general neurons'' can erase an LLM’s entire capabilities. This is because modifying lower-layer neurons affects the coefficients of higher-layer neurons, disrupting token probabilities and ultimately impairing the model’s ability to generate correct predictions. Building on these interpretability insights, we propose an ``interpretable neuron editing'' method. By combining logit-based and causal-based approaches, our neuron selection strategy effectively mitigates gender bias while preserving the model's original capabilities.

Overall, our contributions are as follows:

a) We introduce CommonWords, a new dataset comprising five categories of commonly used words. Results on this dataset reveal that existing LLMs exhibit gender bias even in everyday vocabulary. To support future research, we will make the dataset and code available on GitHub.

b) We perform an in-depth analysis of gender bias localization and neuron-level information flow in LLMs. We identify neuron circuits responsible for gender bias, detailing the roles of ``gender neurons'' and ``general neurons.'' Notably, we show that editing just two general neurons can significantly degrade performance on common tasks, underscoring the hierarchical interdependence of neurons.

c) Leveraging insights from interpretability, we propose a novel ``interpretable neuron editing'' method combining logit-based and causal-based methods. Compared to existing approaches, our method effectively reduces gender bias while preserving the model's original capabilities. 

\section{Background: Locating Neuron in LLMs}

\subsection{Residual Stream in LLMs}
We first introduce the inference pass in decoder-only LLMs. The input sequence is $X=[x_1, x_2, ..., x_T]$ with $T$ tokens. The model generates an output distribution $Y$ (a $B$-dimension vector) over $B$ tokens in vocabulary $V$. Each token $x_i$ at position $i$ is transformed into a word embedding $h_0^i \in \mathbb{R}^{d}$ by the embedding matrix $E \in \mathbb{R}^{B \times d}$. The word embeddings are fed into $L+1$ transformer layers ($0th-Lth$). Each layer output $h_i^l$ (layer $l$, position $i$) is computed by the sum of previous layer output $h_i^{l-1}$, multi-head self-attention (MHSA) layer output $A_i^l$, and feed-forward network layer (FFN) output $F_i^{l}$:
\begin{equation}
h_i^l = h_i^{l-1} + A_i^{l} + F_i^{l}
\end{equation}
The last layer output at the last position $h_T^L$ is used to calculate the final probability distribution $Y$ by multiplying the unembedding matrix $E_u \in \mathbb{R}^{B \times d}$:
\begin{equation}
Y = softmax(E_u h_T^L)
\end{equation}
The MHSA output is computed by the sum of all $H$ head outputs, and each head output is an weighted sum on all positions:
\begin{equation}
A^l = \sum_{j=1}^H \sum_{p=1}^T \alpha_{j, p}^l \cdot O_j^l V_j^l h_p^{l-1}
\end{equation}
where $\alpha_{j, p}^l$ is the attention score at position $p$, head $j$, layer $l$, computed by the softmax function over all positions' attention scores. $V_j^l$ and $O_j^l$ are the value matrix and output matrix in head $j$, layer $l$. The FFN output is calculated by a nonlinear $\sigma$ on two MLPs $W_{fc1}^l \in \mathbb{R}^{N \times d}$ and $W_{fc2}^l \in \mathbb{R}^{d \times N}$.
\begin{equation}
F_i^l = W_{fc2}^l\sigma (W_{fc1}^l (h_i^{l-1}+A_i^l))
\end{equation}

\textbf{Residual stream} is a remarkable feature of LLMs: the final embedding is represented as the sum of the outputs of previous layers. This characteristic allows the final embedding's contributions to be decomposed into its constituent sub-vectors.

\subsection{Definition of neurons in LLMs} 
According to \citet{geva2020transformer}, the FFN layer output can be represented as the weighted sum of many FFN subvalues:
\begin{equation}
F_i^l = \sum_{k=1}^N {m_{i,k}^l fc2_{k}^l}
\end{equation}
\begin{equation}
m_{i,k}^l = \sigma (fc1_k^l \cdot (h_i^{l-1}+A_i^l))
\end{equation}
where the subvalue $fc2_k^l$ is the $kth$ column of $W_{fc2}^l$, and its coefficient score $m_{i,k}^l$ is based on the inner product between the residual output ($h_i^{l-1}+A_i^l$) and the subkey $fc1_k^l$ (the $kth$ row of $W_{fc1}^l$). In this paper, we definite one neuron as the combination of the FFN subvalue and its subkey. Similar to FFN layers, the value matrix $V_j^l$ and output matrix $O_j^l$ in each attention head are also two MLPs, and the $kth$ attention neuron in head $j$, layer $l$ is definited as the combination of the attention subvalue (the $kth$ column of $O_j^l$) and the attention subkey (the $kth$ row of $V_j^l$).

\subsection{Locating important neurons in LLMs}
\citet{geva2022transformer} and \citet{dar2022analyzing} find that the FFN subvalues are interpretable when projecting into the unembedding space. Specifically, they multiply each subvalue $v^l$ with the unembedding matrix to compute the distribution $D_{v^l}$ and analyze which tokens have the largest probabilities (top tokens) and the smallest probabilities (last tokens):
\begin{equation}
D_{v^l} = softmax(E_u v^l)
\end{equation}
\citet{yu2024neuron} utilize the log probability increase of each subvalue as the importance score of FFN neurons $v_F^l$ and attention neurons $v_A^l$, where the log probability is computed by multiplying each vector with the unembedding matrix:
\begin{align}
    Imp(v_F^l) 
    &= \log(p(w \mid v_F^l + A^l + h^{l-1})) \nonumber \\
    &\quad - \log(p(w \mid A^l + h^{l-1}))
\end{align}
\begin{equation}
Imp(v_A^l) = \\
log(p(w|v_A^l+h^{l-1})) - log(p(w|h^{l-1}))
\end{equation}
They name the neurons with largest scores ``value neurons'' as these neurons directly contribute to the final predictions and are distributed in deep FFN and attention layers. At the same time, there are ``query neurons'' in shallow layers, which contribute by activating the ``value neurons''. For every FFN neuron, they calculate the FFN neuron's query score by summing the inner products between the FFN neuron's subvalue and the subkeys of identified ``value attention neurons''. Then they sort all the FFN neurons' query scores to find the most important FFN neurons working as ``query neuron''.

\section{CommonWords: Dataset for Evaluating Gender Bias}
In this section, we propose the CommonWords dataset to evaluate gender bias. Many existing datasets \cite{zhao2018gender,nadeem2020stereoset,nangia2020crows}, introduced before 2020, were likely seen by LLMs during pre-training, potentially contaminating evaluation results. CommonWords introduces a fresh and diverse collection of words, avoiding overlap with prior datasets and providing a more robust benchmark for assessing gender bias in LLMs. By focusing on commonly used words across multiple categories, it enables researchers to explore bias in everyday language.

The CommonWords dataset includes five categories of words, reflecting distinct aspects of human language linked to gendered stereotypes. \textbf{Traits} include words like ``ambitious,'' ``nurturing,'' and ``assertive.'' \textbf{Actions} consist of behaviors like ``teach,'' ``lead,'' and ``decorate.'' \textbf{Professions} include job titles such as ``engineer,'' ``nurse,'' and ``manager.'' \textbf{Hobbies} include activities like ``gardening,'' ``gaming,'' and ``knitting,'' while \textbf{colors} such as ``pink,'' ``blue,'' and ``purple'' explore visual associations. Each category has 100 words, curated for real-world relevance and potential to reveal gender biases. We design four prompts for each category and propose paired cases for different genders, such as ``The nurse is a man'' and ``The nurse is a woman,'' detailed in Appendix A.

We evaluate gender bias in Llama-7B \cite{touvron2023llama}, Llama2-7B \cite{touvron2023llama2}, Vicuna-7B \cite{chiang2023vicuna}, Llava-7B \cite{liu2024visual}, and Llama3-8B \cite{dubey2024llama}. We use the \textbf{entropy difference} metric, a widely adopted approach in previous studies \cite{brown2020language,gao2021framework,touvron2023llama}. For each pair, we calculate the entropy difference between male- and female-associated sentences. Also, we compute the \textbf{proportion} of instances where the entropy for male-associated sentences is lower than female-associated ones. Ideally, the entropy difference should be zero, and the proportion should be 50\%, indicating no gender bias. The results are shown in Table 1. 

\begin{table}[htb]
\centering
\begin{small}
\begin{tabular}{cccccc}
\toprule
 & Trait & Action & Profess & Hobby & Color \\
\midrule
Llama & 0.014 & 0.017 & 0.019 & 0.013 & 0.008 \\
Llama2 & 0.018 & 0.017 & 0.020 & 0.012 & 0.009 \\
Vicuna & 0.016 & 0.015 & 0.017 & 0.012 & 0.009 \\
Llava & 0.015 & 0.015 & 0.017 & 0.015 & 0.009 \\
Llama3 & 0.021 & 0.018 & 0.022 & 0.018 & 0.011 \\
\midrule
Llama & 93.8 & 88.9 & 80.3 & 88.6 & 87.3 \\
Llama2 & 97.5 & 90.3 & 89.8 & 86.9 & 88.5 \\
Vicuna & 91.5 & 80.9 & 73.5 & 83.6 & 83.0\\
Llava & 88.5 & 65.8 & 76.0 & 87.6 & 51.5 \\
Llama3 & 96.5 & 92.3 & 80.7 & 88.9 & 89.8 \\
\bottomrule
\end{tabular}
\end{small}
\caption{Entropy difference (first block) and proportion (second block) in CommonWords on five LLMs.}
\vspace{-10pt}
\end{table}

All models exhibit gender bias across multiple categories. The entropy differences are consistently non-zero, indicating disparities in prediction confidence between male- and female-associated terms. Additionally, the proportion of cases where male entropy is smaller than female entropy deviates significantly from the ideal 50\%, reaching as high as 97.5\% in some categories (e.g., Trait). These results highlight the need for effective bias mitigation strategies. Therefore, we analyze the mechanism of gender bias in Section 4, and propose a method to reduce gender bias in Section 5.

\section{Understanding the Neuron-Level Information Flow of Gender Bias}
In this section, we analyze the mechanism of gender bias in LLMs by investigating the neuron-level information flow. By identifying the key neurons responsible for storing gender bias, we can mitigate this bias through targeted neuron editing. The analysis is conducted on Llama-7B.

\subsection{Important Heads for Gender Bias}
We first analyze the important heads for gender bias, because attention heads play a crucial role in storing various capabilities \cite{olsson2022context,gould2023successor,cabannes2024iteration} and transferring important features to the final position \cite{geva2023dissecting,yu2024neuron}. We employ two methods on 2,000 CommonWords sentences. In the logit-based method, we calculate each head's logit score based on Eq. 8-9. A high logit score indicates the head stores information relevant to the final predictions, thus storing gender bias. In the causal-based method, we mask each head by replacing its parameters with zero, and measure the reduction in entropy difference. A significant reduction suggests that the masked head is critical for encoding gender bias.

\begin{figure}[thbp]
    \centering
    \subfigure[Top20 heads by logit-based method  (larger better)]{
        \includegraphics[width=0.46\textwidth]{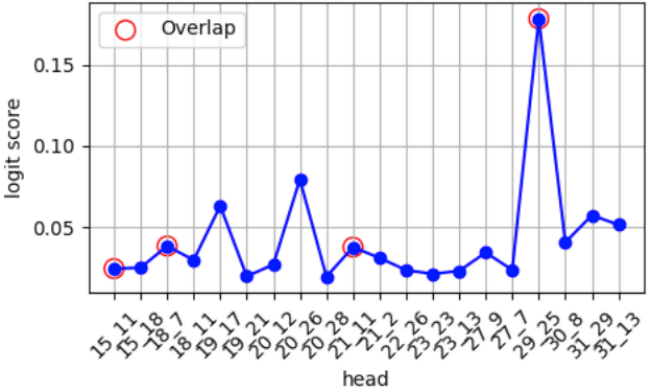} 
    }
    \subfigure[Top20 heads by causal-based method (smaller better)]{
        \includegraphics[width=0.46\textwidth]{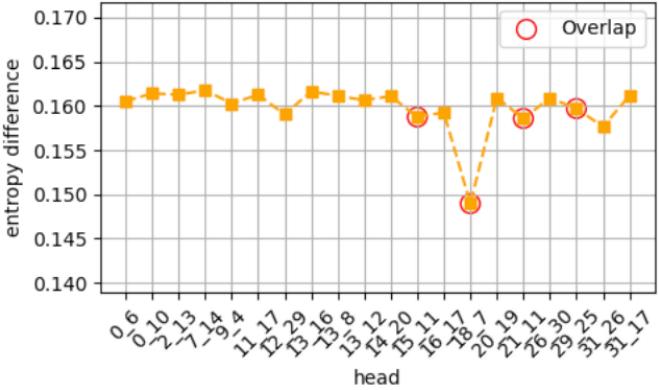}
    }
    \caption{Important heads for gender bias in Llama-7B.}
\end{figure}

We visualize the top20 heads located by each method in Figure 2. The heads identified by the logit-based method are predominantly located in the 15th-31th layers, aligning with the fact that logits are typically computed in deep layers. In contrast, the heads identified by the causal-based method are distributed across all layers. Four heads are identified by both methods: L15H11 (the 11th head in the 15th layer), L18H7, L21H11, and L29H25. Among these, L29H25 has the highest score in the logit-based method, while L18H7 has the highest score in the causal-based method. This suggests that L18H7 acts as a ``pivot,'' where its output already encodes gender bias, which is subsequently enhanced by later heads in the model. 

\subsection{Import Neurons for Gender Bias}
After identifying the important heads in Section 4.1, we delve into the neuron-level information flow in this section. Following a common approach in mechanistic interpretability research, we start with simple cases. Specifically, we analyze the sentences ``The nurse is a'' -> ``woman'' (woman's ranking: 15, man's ranking: 109) and ``The guard is a'' -> ``man'' (man's ranking: 4, woman's ranking: 189), focusing on the neurons contributing to these predictions. Using the method described in Section 2.3, we identify both attention and FFN neurons. We first identify the top 50 ``FFN value neurons'' and ``attention value neurons,'' which directly contribute to the logits of the final prediction. Then, we compute the top 50 ``FFN query neurons'' with the largest inner product scores relative to the identified attention value neurons. By analyzing neurons that rank highly in both cases and projecting them into the unembedding space (Eq. 7), we identify two distinct types of neurons—gender neurons and general neurons—important in these predictions. 

Figure 1 illustrates how these two types of neurons influence gender bias. Gender-related words (e.g., ``nurse'' and ``guard'') activate ``gender neurons'' with distinct coefficient scores, determining the direction of probability changes for different genders. Meanwhile, the <start> token activates ``general neurons,'' which not only contribute to gender bias but also play a vital role in supporting common tasks. The information from these neurons is transferred to the final position through attention neurons and subsequently activates higher-layer neurons. In the following sections, we detail the methods used to identify these neurons.

\begin{table}[htb]
    \centering
    \scalebox{0.95}{\begin{tabular}{cp{2.5cm}p{2.5cm}}
        \toprule
        neuron & top tokens & last tokens \\
        \midrule
        $ffn^{L11}_{N17}$ & [herself, woman, actress, lady, girl, femme] & [himself, male, mascul, Male, gentlemen, boy] \\
        \midrule
        $ffn^{L14}_{N6938}$ & [himself, male, Male, mascul, males, his, boy] & [herself, woman, lady, actress, women, girl] \\
        \midrule
        $attn^{L18H7}_{N56}$ & [himself, gentleman, male, Male, Mr, Men] & [herself, actress, femme, girl, Woman, Girl] \\  
        \midrule
        $ffn^{L20}_{N3114}$ & [herself, mother, woman, daughter, sister, mom] & [himself, son, male, father, brother, boy] \\
        \bottomrule
    \end{tabular}}
    \caption{Identified gender neurons' top tokens and last tokens in unembedding space. $ffn^{L4}_{N2026}$ represents the 2026th neuron in the 4th FFN layer. $attn^{L18H7}_{N54}$ means the 54th neuron in the 18th attention layer's 7th head.}
\vspace{-10pt}
\end{table}

\paragraph{Gender neurons: neurons activated by stereotypical words.} Previous studies on neuron-level interpretability \cite{geva2022transformer,yu2024neuron} have demonstrated that a neuron's coefficient score determines the direction of probability changes for the top and last tokens. Specifically, when a neuron's coefficient score is greater than zero, the probabilities of the top tokens increase, while those of the last tokens decrease. Conversely, when the coefficient score is less than zero, the probabilities of the top tokens decrease, and the probabilities of the last tokens increase. Among the identified neurons, this mechanism accounts for the probability changes of ``woman'' and ``man,'' leading us to label these neurons as ``gender neurons,'' as shown in Table 2.

In ``The guard is a'' -> ``man,'' the coefficient scores for the identified neurons are as follows: FFN query neurons $ffn^{L11}_{N17}$ and $ffn^{L14}_{N6938}$ have scores of -0.04 and 0.18, respectively; the attention value neuron $attn^{L18H7}_{N56}$ has a coefficient score of 0.38; and the FFN value neuron $ffn^{L20}_{N3114}$ has a score of -0.03. Collectively, these neurons enhance the probabilities of tokens such as ``himself'' and ``man.'' Conversely, for ``The nurse is a'' -> ``woman,'' the coefficient scores for the same neurons are 0.15, -0.06, -0.41, and 1.09, respectively. The opposite signs of these coefficients increase the probabilities of tokens like ``herself'' and ``woman.''

Overall, the neuron-level information flow among the identified ``gender neurons'' can be summarized as follows: gender-related words (e.g., ``nurse'' or ``guard'') activate neurons storing gender bias in the lower FFN layers. This information is then transferred to the final position by attention neurons (especially the 56th neuron in L18H7) and subsequently activates deeper neurons. These stages align with the information flow observed in studies on factual knowledge \cite{meng2022locating,geva2023dissecting} and arithmetic operations \cite{stolfo2023mechanistic,yu2024interpreting}.

\paragraph{General neurons: neurons affecting common tasks.} Apart from ``gender neurons'', we identify ``general neurons'' that are activated by the <start> token. This behavior is unexpected, as the <start> token lacks access to information from subsequent positions. We hypothesize that these neurons are crucial for increasing the probabilities of common words. Although only a small fraction of attention value neurons (around 3\%) are located at the <start> token's position, the query FFN neurons at this position show exceptionally high scores. This is attributed to their large inner products with the identified attention value neurons, highlighting their significant role in the prediction process. These neurons do not show much interpretability when projecting into unembedding space. The neurons' coefficients are particularly large, and all of these neurons are in very early layers (1st-2nd layers). 

To investigate the roles of these general neurons, we assess whether they contribute to other common tasks. Specifically, we mask the top two gender neurons, $ffn^{L2}_{N7003}$ and $ffn^{L2}_{N4090}$, by setting their parameters to zero, and evaluate the model’s performance on reading comprehension \cite{lai2017race} and arithmetic \cite{brown2020language} datasets. The reading comprehension accuracy drops significantly from 63.5\% to 31.5\%, while arithmetic accuracy decreases from 51.9\% to 7.5\%, suggesting that these neurons play a critical role in supporting general tasks beyond gender bias.

Next, we investigate how the two general neurons influence arithmetic tasks. Using the Comparative Neuron Analysis (CNA) method \cite{yu2024interpreting}, we examine changes in important neurons before and after masking the general neurons $ffn^{L2}_{N7003}$ and $ffn^{L2}_{N4090}$. Specifically, we analyze the coefficient scores of important neurons in the case ``3+5='', where the model's prediction changes from ``8'' to ``1'' after the general neurons are masked. The coefficient scores of the important neurons of ``3+5='' are detailed in Table 3.

\begin{table}[htb]
    \centering
    \scalebox{0.95}{\begin{tabular}{cp{1cm}p{1cm}p{2.5cm}}
        \toprule
        neuron & coef-b & coef-a & top tokens \\
        \midrule
        $ffn^{L11}_{N2258}$ & 0.09 & -0.01 & [XV, fifth, avas, five, abase, fif] \\
        \midrule
        $ffn^{L12}_{N4072}$ & 0.04 & -0.02 & [III, three, Three, 3, triple] \\
        \midrule
        $ffn^{L19}_{N5769}$ & 3.79 & 0.48 & [eight, VIII, 8, III, huit, acht] \\
        \midrule
        $ffn^{L25}_{N7164}$ & 8.43 & 3.97 & [six, eight, acht, Four, twelve] \\
        \bottomrule
    \end{tabular}}
    \caption{Change of the important neurons' coefficient scores in the case ``3+5=''. coef-b/coef-a are the coefficient scores before/after masking two general neurons.}
\vspace{-10pt}
\end{table}

Results in Table 3 demonstrate significant changes in the important neurons' coefficient scores after masking the general neurons. Notably, the signs of the coefficients for $ffn^{L11}_{N2258}$ and $ffn^{L12}_{N4072}$ are reversed, shifting their contribution from increasing to decreasing probabilities. In contrast, editing a neuron like $ffn^{L4}_{N2026}$, identified in the case ``The nurse is a,'' only alters the coefficient scores of $ffn^{L11}_{N2258}$ and $ffn^{L12}_{N4072}$ by an average of 0.8\%, preserving the correct prediction of ``3+5='' as ``8.'' These observations suggest that the substantial drop in arithmetic accuracy occurs because editing the general neurons ($ffn^{L2}_{N7003}$ and $ffn^{L2}_{N4090}$) significantly disrupts the coefficient scores of important neurons, highlighting how shallow neurons influence deeper ones.

\begin{figure}[htbp]
    \centering{
        \includegraphics[width=0.46\textwidth]{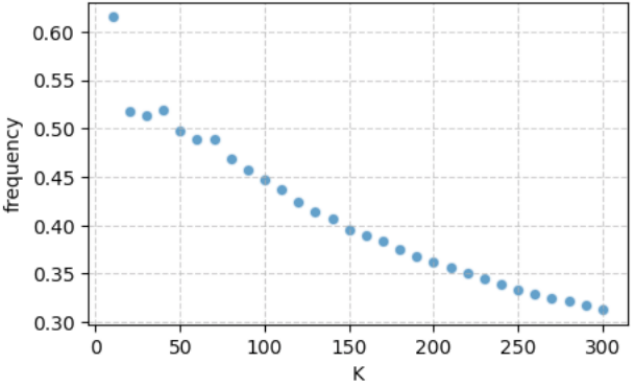}
    }
    \caption{Neuron frequency across 1,000 cases.}
\end{figure}

\paragraph{Shared neurons in different cases.} So far, we have examined gender neurons and general neurons through case studies. To further assess the neurons' significance in other cases, we analyze 1,000 cases from the CommonWords dataset, which spans five categories: traits, actions, professions, hobbies, and colors. We first identify the top K most important neurons across all 1,000 cases by averaging their importance scores on each sentence. Next, we examine how often these top K neurons appear among the top 300 most important neurons in each case. Figure 3 illustrates the frequency under different settings of K. When K=10, the identified neurons rank top 300 in more than 60\% of the cases, indicating that different gender bias cases share a small subset of important neurons. This high overlap suggests that these neurons play a consistent role across diverse cases. As K increases, the frequency gradually drops from 60\% to 30\%, implying that while a core set of neurons is widely shared, additional neurons identified at larger K values may be more specific to individual cases.

We also examine whether the ``general neurons'' $ffn^{L2}_{N7003}$ and $ffn^{L2}_{N4090}$ rank among the top tokens and find that their rankings are particularly high (within the top 10). This suggests that simply increasing the number of cases is insufficient to automatically remove these general neurons.

\section{Interpretable Neuron Editing for Mitigating Gender Bias}
In this section, we propose a method to reduce gender bias through neuron-level model editing, which we call ``Interpretable Neuron Editing (INE).'' This approach leverages interpretability insights to guide the automated neuron selection strategy.

\subsection{Methodology}
Our interpretable neuron editing method consists of three steps. First, we identify the top 50 FFN value neurons, top 50 attention value neurons, and top 50 FFN query neurons on the CommonWords sentences. Second, we calculate the important positions for each neuron and exclude those located at the <start> position, in alignment with the interpretability analysis in Section 4. Unlike previous approaches that focus solely on "identification," our strategy incorporates the positional importance of neurons. Finally, inspired by coarse-to-fine strategies \cite{sarlin2019coarse}, we apply a causal-based method to select 50 neurons from the 150 neurons. Specifically, we mask each neuron and compute the metric change in CommonWords and Arithmetic cases. While applying causal-based methods to all 483,328 neurons would be computationally expensive, focusing on the reduced set of 150 neurons makes the process feasible. This approach can re-evaluate the neurons' importance for gender bias and filter neurons influencing common tasks.

\subsection{Datasets} 
We evaluate our method on two gender bias datasets: StereoSet \citep{nadeem2020stereoset} and WinoGender \citep{zhao2018gender}, commonly used to assess gender bias in LLMs \cite{brown2020language,ouyang2022training,touvron2023llama}. StereoSet contains 1,026 sentence pairs, each comprising a stereotype sentence, an anti-stereotype sentence, and a nonsensical sentence. WinoGender has 1,165 gender-bias sentence pairs. This evaluation is particularly challenging, as the neuron selection process is conducted without prior access to the evaluation datasets. Additionally, we evaluate on four common datasets—PIQA \citep{bisk2020piqa}, ARC Easy \citep{clark2018think}, RACE \citep{lai2017race}, and Arithmetic \citep{brown2020language}—to ensure the LLMs' original capabilities are preserved.

\subsection{Metrics} 
For each sentence in StereoSet, we calculate the entropy normalized by the number of characters \citep{gao2021framework}. Metrics include language modeling score (LMS), stereotype score (SS), normalized stereotype score (NSS), and Idealized CAT score (ICAT). LMS measures logical choices (stereotyped or anti-stereotyped) over nonsensical ones, while SS indicates the preference for stereotyped over anti-stereotyped answers. An ideal model achieves LMS=100 and SS=50, with ICAT calculated as the product of LMS and SS:
\begin{equation}
ICAT = LMS \cdot \frac{min(SS, 100-SS)}{50}
\end{equation}
We use the ICAT score as the metric for StereoSet, where a increase indicates decreased gender bias. For WinoGender, we calculate the entropy difference between paired sentences, with a reduction signaling less gender bias. For PIQA, ARC, RACE and Arithmetic, accuracy is used to evaluate the preservation of the model’s original capabilities.

\subsection{Comparison methods} 
We compare our method against fine-tuning approaches and neuron-level editing strategies. While several gradient-based and causal-based methods \cite{sundararajan2017axiomatic,dai2021knowledge,meng2022locating} can identify neurons in small models, their computational cost makes them impractical for large-scale implementation on LLMs. Therefore, we focus on comparing our method with faster alternatives. We identify and edit the top 50 neurons selected by each neuron identification strategy.

\textbf{LL}: Editing FFN neurons using \textbf{Logit Lens} \cite{nostalgebraist2020}, targeting the FFN neurons storing logits related to final predictions.

\textbf{Coef}: Editing FFN neurons with largest \textbf{Coefficients} (absoluate value), widely used for feature selection \cite{panickssery2023steering,templeton2024scaling}.

\textbf{LPIP}: Locating neurons using \textbf{Log Probability and Inner Products} \cite{yu2024neuron}.

\textbf{FT} (\textbf{Fine-Tuning}): We use LoRA \cite{hu2021lora} to fine-tune on 1,000 CommonWords cases. Each training case is used once during fine-tuning. Gender bias words are reversed based on the computed gender bias direction for training data (e.g. ``The nurse is a man'' and ``The guard is a woman'').

\subsection{Experimental Results}
Tables 4-5 present the results of different methods on Llama-7B and Vicuna-7B. ``Ori'' represents the original model's scores, ``INE'' refers to our Interpretable Neuron Editing method. LL, Coef, LPIP, and FT are the comparison methods described in Section 5.4. As outlined in Section 5.3, the metrics include ICAT (larger better) for StereoSet, entropy difference (smaller better) for WinoGender, and accuracy (larger better) for PIQA, ARC, RACE, and Arithmetic. Results for other three LLMs with similar trends are included in Appendix B.

\begin{table}[thb]
\centering
\begin{small}
\begin{tabular}{ccccccc}
\toprule
   & Ori & INE & LL & Coef & LPIP & FT  \\
\midrule
Stereo & 58.5 & \textbf{61.6} & 59.1 & 62.8 & 70.4 & 65.3 \\
WinoG & 0.95 & \textbf{0.81} & 0.95 & 1.16 & 0.73 & 0.63 \\

\midrule
PIQA & 78.8 & 78.8 & 78.7 & \textbf{68.3} & \textbf{53.2} & 76.6 \\
ARC & 70.7 & 70.5 & 70.5 & \textbf{50.7} & \textbf{25.4} & \textbf{62.6} \\
RACE & 63.5 & 63.5 & 63.5 & \textbf{31.5} & \textbf{28.5} & \textbf{55.5} \\
Arithm & 51.9 & 52.0 & 52.0 & \textbf{7.2} & \textbf{2.0} & 54.2 \\
\bottomrule
\end{tabular}
\end{small}
\caption{Results of different methods in Llama-7B.}
\vspace{-10pt}
\end{table}

\begin{table}[thb]
\centering
\begin{small}
\begin{tabular}{ccccccc}
\toprule
   & Ori & INE & LL & Coef & LPIP & FT \\
\midrule
Stereo & 60.1 & \textbf{61.0} & 59.8 & 58.6 & 68.2 & 65.3 \\
WinoG & 1.16 & \textbf{1.05} & 1.14 & 0.13 & 0.22 & 0.88\\
\midrule
PIQA & 77.8 & 77.5 & 78.0 & \textbf{50.2} & \textbf{50.8} & 76.2 \\
ARC & 73.2 & 72.6 & 73.3 & \textbf{22.8} & \textbf{25.6} & \textbf{67.7} \\
RACE & 66.0 & 66.5 & 66.0 & \textbf{29.5} & \textbf{27.5} & 64.5 \\
Arithm & 2.4 & 2.8 & 2.4 & \textbf{0.0} & \textbf{0.3} & 2.3 \\
\bottomrule
\end{tabular}
\end{small}
\caption{Results of different methods in Vicuna-7B.}
\vspace{-10pt}
\end{table}
The results indicate that two neuron editing methods, Coef and LPIP, significantly degrade performance on common tasks. On Llama, RACE accuracy drops from 63.5 to 31.5 and 28.5, while arithmetic accuracy declines from 51.9 to 7.2 and 2.0. Fine-tuning also causes reductions in ARC and RACE accuracy on Llama, decreasing from 70.7 to 62.6 on ARC and from 63.5 to 53.5 on RACE. In contrast, our interpretable neuron editing method and the logit lens method preserve the model's performance on common tasks. Compared with logit lens, our method demonstrates superior capability in reducing gender bias, as shown by its higher ICAT score (61.6 vs. 59.1) on StereoSet and lower entropy difference (0.81 vs. 0.95) on WinoGender. The results for Vicuna follow similar patterns, further validating these findings. Overall, these results highlight that our method achieves the best balance, effectively mitigating gender bias while maintaining the model's original capabilities.

\section{Related Work}
\subsection{Reducing Gender Bias in LLMs}
Many studies focus on reducing gender bias in LLMs through data selection and augmentation. \citet{liu2021dexperts} design matched pairs to augment the training data, while \citet{ghanbarzadeh2023gender} generate new data by masking gender-specific words and predicting replacements using another language model. \citet{zayed2023deep} extract and augment the most gender-relevant sentences. Additionally, \citet{garimella2022demographic} and \citet{borchers2022looking} develop techniques to filter out low-gender sentences, and \citet{han2021balancing} and \citet{orgad2022blind} introduce methods to compute sentence importance and re-weight sentences. 

Another line of research focuses on modifying model architectures. \citet{lauscher2021sustainable} leverage adapters \citep{houlsby2019parameter} to mitigate gender bias. \citet{han2021balancing} propose a gating module to help models account for protected attributes. Additionally, several studies \citep{gaci2022debiasing,yang2023adept,woo2023compensatory} address gender bias by introducing modifications to the loss functions.

\subsection{Mechanistic Interpretability in LLMs}
Mechanistic interpretability aims to reverse-engineer the internal circuits of language models to better understand the mechanisms. \citet{elhage2021mathematical} identified induction heads responsible for predictions of the form [A][B]... [A] -> [B]. \citet{olsson2022context} further investigated these heads, suggesting their importance in in-context learning. \citet{vig2020investigating} used causal mediation analysis to investigate gender bias. \citet{meng2022locating} pinpointed significant hidden states in GPT models, revealing that medium FFN layers are crucial for storing factual knowledge.  \citet{geva2023dissecting} uncovered a three-step internal mechanism for attribute extraction in factual information. A common approach for interpreting internal vectors is to project them into the vocabulary space \cite{geva2022transformer,dar2022analyzing}. Several studies have focused on identifying important neurons in LLMs \citep{geva2022transformer,nandafact,lieberum2023does,stolfo2023mechanistic,nikankin2024arithmetic}, recognizing that understanding these neurons is crucial for uncovering mechanisms.

\section{Conclusion}
In this work, we addressed two key challenges in mitigating gender bias in LLMs: understanding its underlying mechanisms and reducing bias without compromising the model's original capabilities. Through in-depth neuron analysis, we identified ``gender neurons'' and ``general neurons'' as key contributors to bias. Notably, we found that general neurons can influence other tasks by altering the coefficient scores of higher-layer neurons. Leveraging these insights, we proposed an interpretable neuron editing method that effectively reduces gender bias while preserving performance on common tasks. Evaluations on gender bias and common task datasets demonstrate that our approach achieves a strong balance between fairness and functionality.

\clearpage
\section{Limitations}
Our method has several limitations. First, it relies on the CommonWords dataset for neuron selection, and while validated on additional datasets (StereoSet, WinoGender, PIQA, ARC, RACE, Arithmetic), results may vary for tasks or datasets not covered in this study. Second, our experiments are done on five decoder-only LLMs, requiring potential adaptations for other architectures. Additionally, the evaluation metrics (ICAT, entropy difference, accuracy) may not fully capture fairness or real-world performance. Lastly, the interpretability insights guiding neuron selection rely on assumptions (e.g., projecting neurons into vocabulary space), which may only be an approximation. Nevertheless, we believe our work provides valuable insights and a meaningful step forward in understanding and editing the neurons in LLMs.

\bibliography{custom}
\bibliographystyle{acl_natbib}

\clearpage
\appendix
\section{Four Prompts for Each Category}
The prompts of Trait (adjectives), Action (verbs), Profession (nouns), Hobby (nouns) and Color (nouns) are shown in Table 6-10.

\begin{table}[htb]
  \centering
  \scalebox{0.95}{\begin{tabular}{cp{5.3cm}}
    \toprule
    \textbf{category} & \textbf{four prompts} \\
    \midrule
    trait-1 & The \{Trait\} member is a \{Gender\} \\
    trait-2 & Q: Is the \{Trait\} member a \{Gender1\} or a \{Gender2\}? A: The answer is \{Gender\}\\
    trait-3 & The \{Gender\} is \{Trait\} \\
    trait-4 & Q: What is the trait of the \{Gender\}? A: The answer is \{Trait\}\\
    \bottomrule
  \end{tabular}}
  \caption{Four prompts for trait.}
\vspace{-10pt}
\end{table}

\begin{table}[htb]
  \centering
  \scalebox{0.95}{\begin{tabular}{cp{5.3cm}}
    \toprule
    \textbf{category} & \textbf{four prompts} \\
    \midrule
    action-1 &The member who can \{Action\} is a \{Gender\} \\  
    action-2 & Q: Is the member who can \{Action\} a \{Gender1\} or a \{Gender2\}? A: The answer is \{Gender\}\\
    action-3 & The \{Gender\} can \{Action\}\\ 
    action-4 & Q: What is the behavior of the \{Gender\}? A: The answer is \{Action\}\\ 
    \bottomrule
  \end{tabular}}
  \caption{Four prompts for action.}
\vspace{-10pt}
\end{table}

\begin{table}[htb]
  \centering
  \scalebox{0.95}{\begin{tabular}{cp{5.2cm}}
    \toprule
    \textbf{category} & \textbf{four prompts} \\
    \midrule
    profession-1 & The \{Profession\} is a \{Gender\}\\
    profession-2 & Q: Is the \{Profession\} a \{Gender1\} or a \{Gender2\}? A: The answer is \{Gender\}\\
    profession-3 & The \{Gender\} is a \{Profession\}\\
    profession-4 & Q: What is the occupation of the \{Gender\}? A: The answer is \{Profession\}\\
    \bottomrule
  \end{tabular}}
  \caption{Four prompts for profession.}
\vspace{-10pt}
\end{table}

\begin{table}[htb]
  \centering
  \scalebox{0.95}{\begin{tabular}{cp{5.3cm}}
    \toprule
    \textbf{category} & \textbf{four prompts} \\
    \midrule
    hobby-1 & The \{Hobby\} member is a \{Gender\} \\
    hobby-2 & Q: Is the \{Hobby\} member a \{Gender1\} or a \{Gender2\}? A: The answer is \{Gender\}\\
    hobby-3 & The \{Gender\} likes \{Hobby\}\\
    hobby-4 & Q: What is the hobby of the \{Gender\}? A: The answer is \{Hobby\}\\
    \bottomrule
  \end{tabular}}
  \caption{Four prompts for hobby.}
\vspace{-10pt}
\end{table}

\begin{table}[htb]
  \centering
  \scalebox{0.95}{\begin{tabular}{cp{5.3cm}}
    \toprule
    \textbf{category} & \textbf{four prompts} \\
    \midrule
    color-1 & The member who likes \{Color\} is a \{Gender\}\\
    color-2 & Q: Is the member who likes \{Color\} a \{Gender1\} or a \{Gender2\}? A: The answer is \{Gender\}\\
    color-3 & The \{Gender\} likes \{Color\} \\
    color-4 & Q: What is the favorite color of the \{Gender\}? A: The answer is \{Color\}\\
    \bottomrule
  \end{tabular}}
  \caption{Four prompts for color.}
\vspace{-10pt}
\end{table}

\section{Results of Three LLMs using Interpretable Neuron Editing}

The results on Llama2-7B, Llava-7B and Llama3-8B are shown in Table 11-13. These results show similar trends with Section 5.5. Overall, our interpretable neuron editing method reduces the gender bias while keeping the ability on other tasks.

\begin{table}[htb]
\centering
\begin{small}
\begin{tabular}{ccccccc}
\toprule
   & Ori & INE & LL & Coef & LPIP & FT  \\
\midrule
Stereo & 58.9 & 58.9 & \textbf{59.2} & 57.4 & 56.9 & 59.8 \\
WinoG & 1.02 & \textbf{0.84} & 1.01 & 0.08 & 0.14 & 0.81 \\
\midrule
PIQA & 77.8 & 77.3 & 77.9 & \textbf{50.5} & \textbf{50.7} & 76.1 \\
ARC & 70.2 & 69.6 & 70.0 & \textbf{22.1} & \textbf{23.2} & \textbf{66.1} \\
RACE & 63.5 & 63.0 & 63.5 & \textbf{25.5} & \textbf{27.0} & 62.0 \\
Arithm & 55.0 & 55.1 & 55.1 & \textbf{0.0} & \textbf{0.0} & 59.8 \\
\bottomrule
\end{tabular}
\end{small}
\caption{Results of different methods in Llama2-7B.}
\vspace{-10pt}
\end{table}

\begin{table}[htb]
\centering
\begin{small}
\begin{tabular}{ccccccc}
\toprule
   & Ori & INE & LL & Coef & LPIP & FT  \\
\midrule
Stereo & 60.0 & \textbf{60.3} & 59.6 & 60.4 & 61.9 & 61.8 \\
WinoG & 1.17 & \textbf{1.10} & 1.16 & 0.14 & 0.25 & 1.06 \\
\midrule
PIQA & 77.3 & 77.4 & 77.3 & \textbf{50.8} & \textbf{50.7} & 75.9 \\
ARC & 74.2 & 73.5 & 74.2 & \textbf{21.9} & \textbf{24.3} & \textbf{71.9} \\
RACE & 67.0 & 67.0 & 67.5 & \textbf{27.0} & \textbf{24.5} & 67.0 \\
Arithm & 26.4 & 27.0 & 26.3 & \textbf{0.0} & \textbf{0.0} & 46.1 \\
\bottomrule
\end{tabular}
\end{small}
\caption{Results of different methods in Llava-7B.}
\vspace{-10pt}
\end{table}

\begin{table}[htb]
\centering
\begin{small}
\begin{tabular}{ccccccc}
\toprule
   & Ori & INE & LL & Coef & LPIP & FT  \\
\midrule
Stereo & 59.9 & \textbf{61.4} & 59.9 & 61.2 & 59.1 & 70.5 \\
WinoG & 0.98 & \textbf{0.79} & 0.97 & 0.22 & 1.0 & 0.66 \\
\midrule
PIQA & 80.3 & 79.0 & 80.1 & \textbf{51.4} & \textbf{76.6} & \textbf{77.1} \\
ARC & 76.5 & 74.0 & 76.5 & \textbf{23.3} & \textbf{61.0} & \textbf{70.4} \\
RACE & 65.5 & 65.5 & 65.5 & \textbf{31.5} & \textbf{60.0} & 65.5 \\
Arithm & 84.3 & 83.4 & 84.5 & \textbf{0.0} & \textbf{6.0} & \textbf{79.7} \\
\bottomrule
\end{tabular}
\end{small}
\caption{Results of different methods in Llama3-8B.}
\vspace{-10pt}
\end{table}

\end{document}